\documentclass[conference]{IEEEtran}
\IEEEoverridecommandlockouts

\usepackage{cite}
\usepackage{amsmath,amssymb,amsfonts, pifont}
\usepackage{algorithmic}
\usepackage{array}
\usepackage{amssymb}
\usepackage{bm}
\usepackage{booktabs}
\usepackage{graphicx}
\usepackage{multirow}
\usepackage{textcomp}
\usepackage{boldline}

\usepackage[table]{xcolor}
\usepackage{caption}
\definecolor{green}{HTML}{2eb04b}
\definecolor{cellred}{RGB}{255, 204, 201}  
\definecolor{cellorange}{RGB}{255, 228, 207} 
\definecolor{cellyellow}{RGB}{254, 255, 211} 

\definecolor{textred}{RGB}{255, 204, 201}  
\definecolor{textorange}{RGB}{255, 228, 207} 
\definecolor{textyellow}{RGB}{254, 255, 211} 

\def\BibTeX{{\rm B\kern-.05em{\sc i\kern-.025em b}\kern-.08em
    T\kern-.1667em\lower.7ex\hbox{E}\kern-.125emX}}

\title{See In Detail: Enhancing Sparse-view 3D Gaussian Splatting with Local Depth and Semantic Regularization \\
}

\author{\IEEEauthorblockN{Zongqi He\textsuperscript{\textsection}, Zhe Xiao\textsuperscript{\textsection}, Kin-Chung Chan, Yushen Zuo, Jun Xiao\textsuperscript{*}, Kin-Man Lam }
\IEEEauthorblockA{\textit{Department of Electrical and Electronic Engineering, The Hong Kong Polytechnic University}\\
Email: \{plume.he, xiao-zhe.xiao, alfred.chen, jun.xiao\}@connect.polyu.hk, \{yushen.zuo, kin.man.lam\}@polyu.edu.hk}
}
\begin{document}

\maketitle
\begingroup\renewcommand\thefootnote{\textsection}
\footnotetext{Equal contribution}
\renewcommand\thefootnote{*}
\footnotetext{Corresponding author}

\begin{abstract}

3D Gaussian Splatting (3DGS) has shown remarkable performance in novel view synthesis. However, its rendering quality deteriorates with sparse inphut views, leading to distorted content and reduced details. This limitation hinders its practical application. To address this issue, we propose a sparse-view 3DGS method. Given the inherently ill-posed nature of sparse-view rendering, incorporating prior information is crucial. We propose a semantic regularization technique, using features extracted from the pretrained DINO-ViT model, to ensure multi-view semantic consistency. Additionally, we propose local depth regularization, which constrains depth values to improve generalization on unseen views. Our method outperforms state-of-the-art novel view synthesis approaches, achieving up to 0.4dB improvement in terms of PSNR on the LLFF dataset, with reduced distortion and enhanced visual quality.

\end{abstract}  

\begin{IEEEkeywords}
Novel view synthesis, Sparse-view rendering, Gaussian Splatting
\end{IEEEkeywords}

\section{Introduction}

Given a set of images captured from different known viewpoints, the goal of novel view synthesis (NVS) is to generate realistic images of the same scene from unseen perspectives while preserving multi-view consistency. NVS techniques are crucial for understanding the 3D world and hold significant industrial value in practical applications across computer vision
\cite{poppe2010survey,xiao2023online,gavrila19963,zuo2024towards,junejo2008cross,zhang2024integrally,seitz2006comparison,furukawa2009accurate,choy20163d,zhang2024structured,fan2017point}, 
graphics \cite{xiao2021feature,li2024hierarchical,xiao2024towards, remondino2006image,xiao2021self, zwicker2002pointshop,xiao2024deep}, and robotics \cite{varley2017shape}.


Neural Radiance Field (NeRF)-based methods \cite{mildenhall2021nerf, barron2021mip, barron2022mip, zhang2020nerf++, xu2022point, liu2020neural, sitzmann2021light} and 3D Gaussian splatting-based methods \cite{kerbl3Dgaussians, chung2023luciddreamer,yan2024multi,zhang2024pixel,lee2024compact,lu2024scaffold} are two leading approaches that have shown remarkable performance in recent years. However, these two approaches typically require input images captured from dense views to produce high-quality images of unseen viewpoints, a condition often difficult to meet in real-world scenarios. As the number of input views decreases, rendering quality inevitably degrades. Currently, generating novel views from sparse inputs remains a significant challenge.


In recent years, several promising methods have been proposed for generating high-quality 3D scenes based on sparse inputs. For instance, RegNeRF \cite{niemeyer2022regnerf} introduces a depth smoothness technique to improve the accuracy of the geometric properties of reconstructed scenes, while DietNeRF \cite{jain2021putting} enhances semantic consistency by encouraging views, encoded by a pretrained CLIP \cite{radford2021learning} vision transformer (ViT), to be closer to each other in the latent space. SparseNeRF \cite{wang2023sparsenerf} leverages dense depth maps, estimated by a pretrained dense prediction transformer (DPT) \cite{ranftl2021vision}, to distill local depth ranking priors, promoting spatial continuity. Although NeRF-based methods have demonstrated promising performance for novel view synthesis from sparse inputs, their slow inference speed and high computational demands limit their applications in real-time products. Recently, 3D Gaussian splatting(3DGS) \cite{kerbl3Dgaussians} has proven effective for real-time, high-quality rendering of 3D scenes, but its potential for novel view synthesis with sparse inputs is not well explored.


\begin{figure}[t]
    \centering
    \includegraphics[width=1\linewidth]{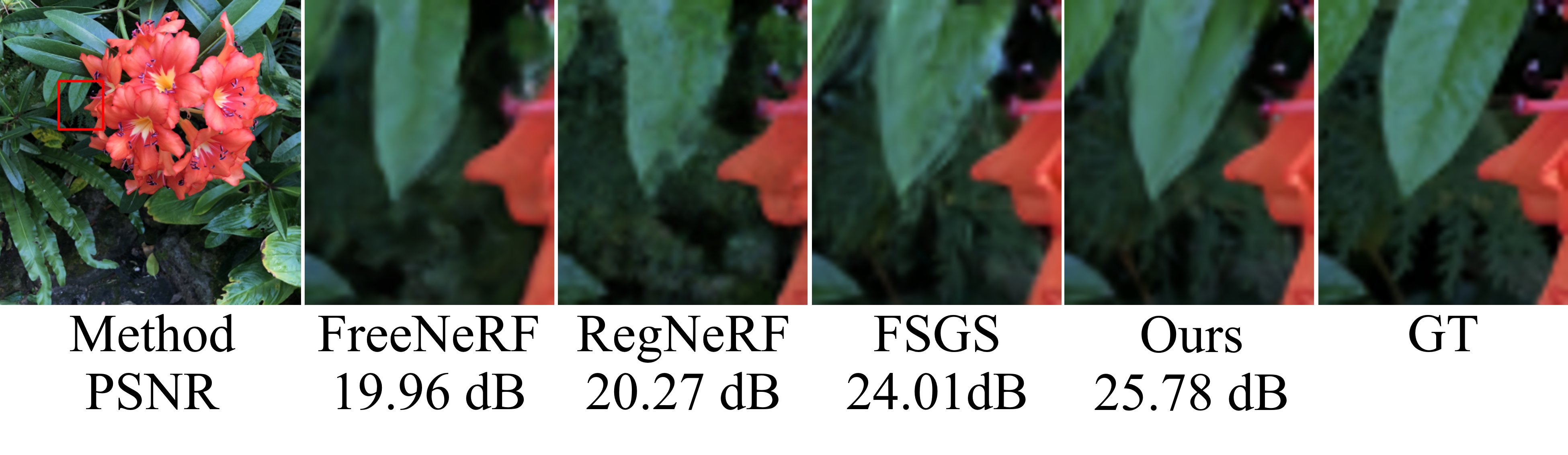}
    \caption{Visual results of FreeNeRF \cite{Yang2023FreeNeRF}, RegNeRF \cite{niemeyer2022regnerf}, FSGS \cite{zhu2023FSGS}, and our SIDGaussian.}
    \label{fig:intro}
\end{figure}

In this paper, we propose a 3DGS method from sparse inputs, named SIDGaussian, which effectively generates finer details in rendered images and preserves their multi-view consistency. Since producing high-quality 3D scenes based on sparse inputs is an inherently ill-posed problem, incorporating prior information is crucial for enhancing performance. To better address this issue, we propose a semantic regularization technique that ensures multi-view semantic consistency, by minimizing the distance between the semantic features of images rendered from training views and side views,  in the latent space, which are extracted using DINO-ViT \cite{zhang2022dino}.

\begin{figure*}[!t]
    \centering
    \includegraphics[width=1\linewidth]{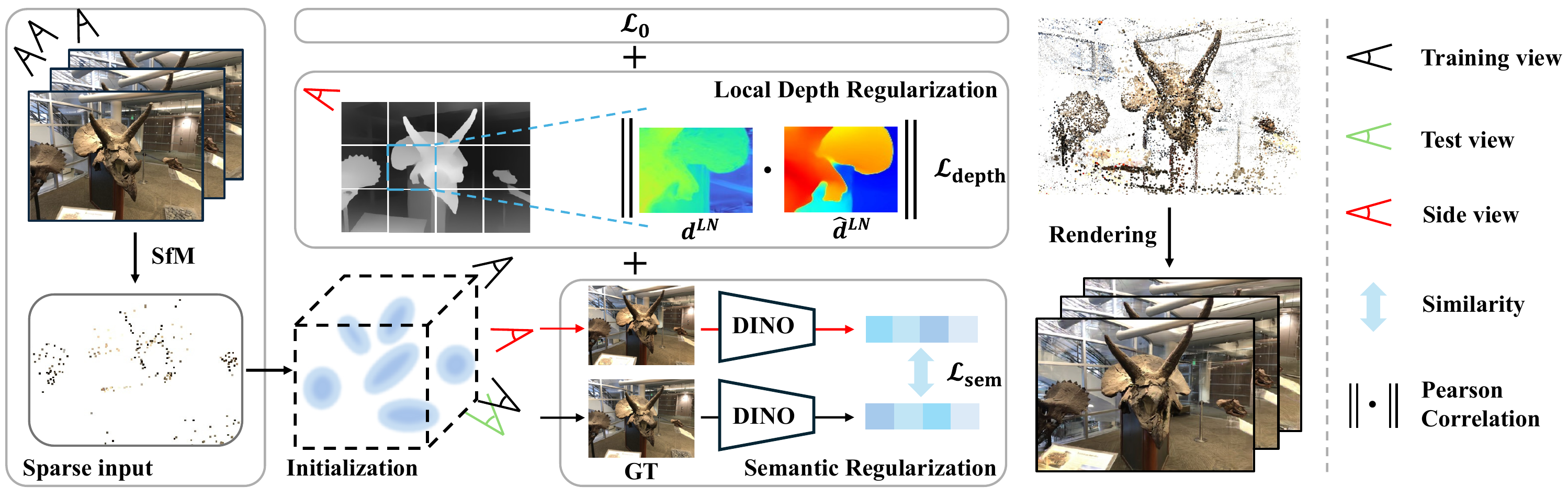}
    \caption{The overall pipeline of our proposed SIDGaussian. 
    A sparse point cloud for the 3D Gaussian initialization is generated from sparse views using SfM. In addition to the $\mathcal{L}_{0}$ loss, the proposed local depth regularization $\mathcal{L}_\text{{depth}}$, and semantic regularization $\mathcal{L}_\text{{sem}}$ are applied during training.
    }
    \label{fig:pipeline}
\end{figure*}

The setting of sparse inputs often results in insufficient appearance coverage and limited geometric information, leading to distorted content in existing methods (see Fig.~\ref{fig:intro}). 
To handle this problem, we further propose a local depth regularization method. Instead of directly using dense depth maps like DDP-NeRF \cite{roessle2022dense}, we locally normalize the depth map and compute the Pearson correlation \cite{cohen2009pearson} between the rendered depth and depth maps from DPT, within local regions. This approach effectively enhances the local geometries of the scene and improves multi-view consistency in the generated outputs.

The main contributions of this paper are summarized as follows:
\begin{itemize}
    \item We propose a 3DGS method, namely SIDGaussian, for novel view synthesis based on sparse inputs, which can achieve real-time and high-quality rendering of 3D scenes.
    \item To ensure multi-view consistency, we introduce a semantic regularization technique that maintains the semantic coherence of rendered images across different viewpoints. Additionally, we propose local depth regularization to reduce distortions and enhance detailed geometries of the scene.
    \item Experiments demonstrate that our method significantly outperforms state-of-the-art novel view synthesis methods, delivering up to a 0.4dB improvement in terms of PSNR on the LLFF dataset. Our method can effectively preserve multi-view consistency and produce visually superior results with minimal distortion.

\end{itemize}

\section{Methodology}
\subsection{Preliminary}

3D Gaussian Splatting is a promising real-time rendering approach for novel view synthesis of 3D scenes. 
It utilizes a set of 3D Gaussians and projects them onto the 2D image plane based on depth information during the splatting process. $\alpha$-blending is crucial for rendering the color $\mathbf{C}_{p}$ of the pixel $p$, as follows:
\begin{equation}
    \mathbf{C}_{p} = \sum_{i=1}^{N}c_{i}\alpha_{i}\prod_{j=1}^{i-1}(1-\alpha_{j}),
\end{equation}
where $c_{i}$ and $\alpha_{i}$ are the color coefficient and the blending weight of the $i$-th 3D Gaussian, respectively. $N$ denotes the total number of 3D Gaussians. The synthesized performance of 3DGS inevitably deteriorates when the number of views decreases. To address this issue, previous work \cite{zhu2023FSGS} additionally leverages global depth information during training, and the loss function is formulated as follows:  
\begin{equation}
    \label{equ:l0}
    \mathcal{L}_{\text{0}} = \lambda \Arrowvert C - \hat{C} \Arrowvert_{1} + \gamma \mathcal{L}_\text{{D-SSIM}}(C, \hat{C}) +  \beta \Arrowvert d(D, \hat{D}) \Arrowvert_{1},
\end{equation}
where $C$ and $\hat{C}$ represent the color values of the ground-truth and estimated images, respectively. $D$ and $\hat{D}$ are the global depth map of the ground-truth and estimated images, respectively, and $d(\cdot)$ is the distance function.

However, solely considering global depth information is not sufficient and often results in distorted content in synthesized images, as shown in Fig.~\ref{fig:intro}. In this paper, we propose semantic regularization and local depth regularization during training, which can effectively enhance the multi-view consistency of synthesized images from different viewpoints and improve detailed content.

\subsection{Proposed Method} 

The pipeline of the proposed method is illustrated in Fig. \ref{fig:pipeline}. 
Specifically, we construct sparse point clouds, using Structure from Motion (SfM), for 3D Gaussian initialization, and perform per-scene optimization, supervised by $\mathcal{L}_\text{0}$ loss, the proposed semantic regularization and local depth regularization during training.

\color{black}

\begin{table*}[!ht]
    \tabcolsep=0.35cm
    \renewcommand\arraystretch{1.5}
    \centering
    \caption{The average PSNR, SSIM, and LPIPS of different methods on the LLFF datasets with 1/8 and 1/4 resolution. The best, second-best, and third-best results are highlighted in red, orange, and yellow, respectively.}
    \begin{tabular}{|c|cccc|cccc|}
    \hline
    \multirow{2}{*}{Methods} & \multicolumn{3}{c|}{1/8 Resolution} & \multicolumn{3}{c|}{1/4 Resolution} \\ \cline{2-7} 
    & \multicolumn{1}{c|}{PSNR$\uparrow$}  & \multicolumn{1}{c|}{SSIM$\uparrow$}  & \multicolumn{1}{c|}{LPIPS$\downarrow$} & \multicolumn{1}{c|}{PSNR$\uparrow$}  & \multicolumn{1}{c|}{SSIM$\uparrow$}  & \multicolumn{1}{c|}{LPIPS$\downarrow$} \\ \hline
    Mip-NeRF \cite{barron2021mip} & \multicolumn{1}{c|}{16.11} & \multicolumn{1}{c|}{0.401} & \multicolumn{1}{c|}{0.460} & \multicolumn{1}{c|}{15.22} & \multicolumn{1}{c|}{0.351} & \multicolumn{1}{c|}{0.540} \\ \hline
    
    3DGS \cite{kerbl3Dgaussians}  & \multicolumn{1}{c|}{17.43} & \multicolumn{1}{c|}{0.522} & \multicolumn{1}{c|}{0.321} & \multicolumn{1}{c|}{16.94} & \multicolumn{1}{c|}{0.488} & \multicolumn{1}{c|}{0.402} \\ \hline
    \hlineB{2}
    DietNeRF \cite{jain2021putting}  & \multicolumn{1}{c|}{14.94} & \multicolumn{1}{c|}{0.370} & \multicolumn{1}{c|}{0.496} & \multicolumn{1}{c|}{13.86} & \multicolumn{1}{c|}{0.305} & \multicolumn{1}{c|}{0.578} \\ \hline
    
    RegNeRF \cite{niemeyer2022regnerf} & \multicolumn{1}{c|}{19.08} & \multicolumn{1}{c|}{0.587} & \multicolumn{1}{c|}{0.336} & \multicolumn{1}{c|}{18.06} & \multicolumn{1}{c|}{0.535} & \multicolumn{1}{c|}{0.411} \\ \hline
    
    FreeNeRF \cite{Yang2023FreeNeRF} & \multicolumn{1}{c|}{19.63} & \multicolumn{1}{c|}{0.612} & \multicolumn{1}{c|}{\cellcolor{cellyellow}{0.308}} & \multicolumn{1}{c|}{18.73} & \multicolumn{1}{c|}{0.562} & \multicolumn{1}{c|}{\cellcolor{cellyellow}{0.384}} \\ \hline
    
    SparseNeRF \cite{wang2023sparsenerf} & \multicolumn{1}{c|}{\cellcolor{cellyellow}{19.86}} & \multicolumn{1}{c|}{\cellcolor{cellyellow}{0.624}} & \multicolumn{1}{c|}{0.328}& \multicolumn{1}{c|}{\cellcolor{cellyellow}{19.07}} & \multicolumn{1}{c|}{\cellcolor{cellyellow}{0.564}} & \multicolumn{1}{c|}{0.401} \\ \hline
    
    FSGS \cite{zhu2023FSGS} & \multicolumn{1}{c|}{\cellcolor{cellorange}{20.31}} & \multicolumn{1}{c|}{\cellcolor{cellorange}{0.652}} & \multicolumn{1}{c|}{\cellcolor{cellorange}{0.288}} & \multicolumn{1}{c|}{\cellcolor{cellorange}{19.88}} & \multicolumn{1}{c|}{\cellcolor{cellorange}{0.612}} & \multicolumn{1}{c|}{\cellcolor{cellorange}{0.340}} \\ \hline
    
    \textbf{SIDGaussian (ours)} & \multicolumn{1}{c|}{\cellcolor{cellred}{20.71}} & \multicolumn{1}{c|}{\cellcolor{cellred}{0.708}} & \multicolumn{1}{c|}{\cellcolor{cellred}{0.205}}    
     & \multicolumn{1   }{c|}{\cellcolor{cellred}{20.02}}      & \multicolumn{1}{c|}{\cellcolor{cellred}{0.667}}      &    \multicolumn{1}{c|}{\cellcolor{cellred}{0.284}}   \\ \hline
    \end{tabular}
    \label{tb:LLFF}
\end{table*}

\begin{figure*}[!ht]
    \centering
    \includegraphics[width=1\linewidth]{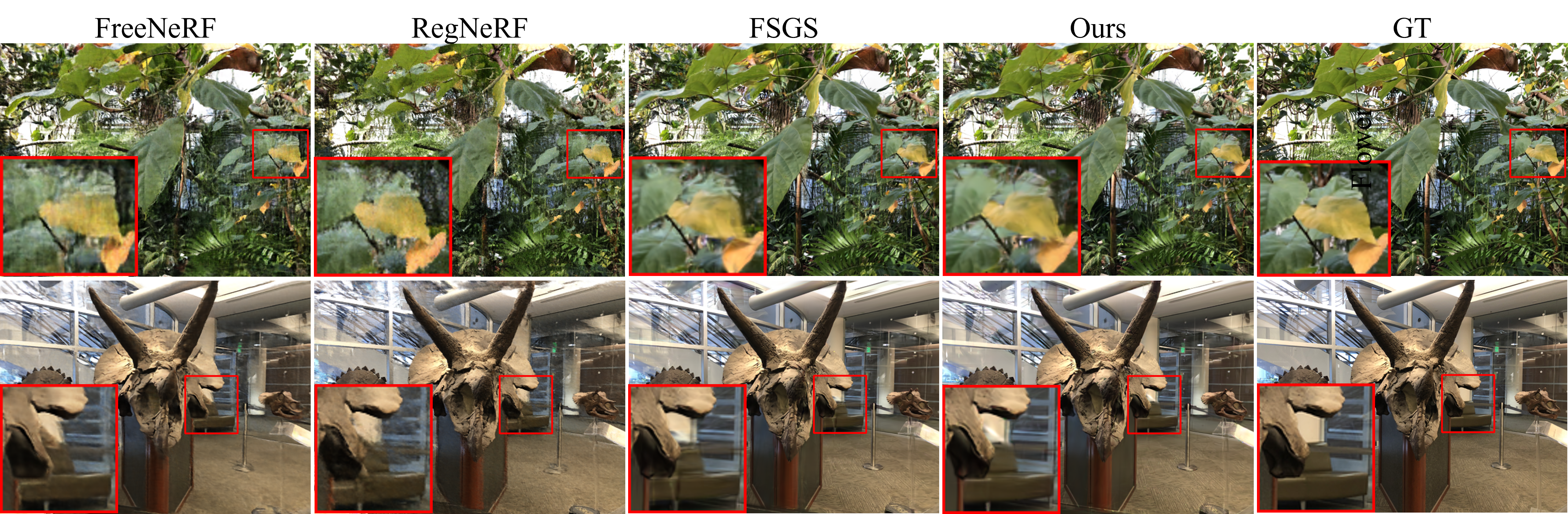}
    \caption{Visual results of the scenes ``Leaves" and ``Horns" generated by FreeNeRF \cite{Yang2023FreeNeRF}, RegNeRF \cite{niemeyer2022regnerf}, FSGS \cite{zhu2023FSGS}, and our SIDGaussian.}
    \label{fig:LLFF}
\end{figure*}

\subsubsection{Semantic Regularization}

Previous works \cite{xiao2024deep1, jain2021putting,xie2023structure, xu2022sinnerf, xie2024satr,chan2024point, xiao2024dmf} have proven the efficacy of semantic regularization in reconstructing the global structure of the scenes. This regularization encourages side views, unseen from training views, to reconstruct contents with similar semantic meaning to the training views. In our method, we propose to enhance optimization with a similar strategy. Specifically, we generate side views, using the same approach described in \cite{zhu2023FSGS}, and adopt a pretrained DINO-ViT \cite{zhang2022dino} to encode both side views and training views. We formulate the semantic regularization term as the distance between these two embeddings, as follows:

\begin{equation}
    \mathcal{L}_\text{{sem}} = \parallel {f}_{vit}(P') - {f}_{vit}(P)\parallel^{2},
\end{equation}
where ${f}_{vit}$ refers to the DINO encoder and $\Vert \Vert^{2}$ denotes $L_{2}$ distance. $P'$ and $P$ refer to the patches randomly cropped from rendered images of the side views and training views, respectively.

\subsubsection{Local Depth Regularization}

Previous work \cite{zhu2023FSGS} utilizes global depth information to facilitate the reconstruction of the 3D geometry. However, this method struggles to improve scenes containing various objects at multiple scales. Particularly, the use of global depth information tends to focus on global features while compressing and disregarding the intricate details of depth information.
To address this issue, we propose to enhance the local details of 3D objects by introducing local depth regularization on side views, alongside global depth regularization. Specifically, we locally normalize depth maps and encourage the similarity between local patches $\mathcal{P}_r$, obtained from depth rendering, and $\mathcal{P}_t$, processed by a pretrained Dense Prediction Transformer (DPT) \cite{ranftl2021vision}.
The normalization process is formulated as follows:
\begin{equation}
    d^{LN}(x) = \frac{d(x) - \mu}{\sigma + \epsilon}, \quad \text{s.t. } x \in \mathcal{P},
\end{equation}
where $\mathcal{P} \subseteq \{\mathcal{P}_r, \mathcal{P}_t\}$, $\mu$ and $\sigma$ are the mean and standard deviation of the corresponding local patch, respectively, and $\epsilon$ is a value for numerical stability.
In addition, we utilize a soft metric, Pearson correlation \cite{cohen2009pearson}, to measure the similarity of two depth patches while mitigating the scale ambiguity issue mentioned in \cite{chan2024point}. The proposed local depth regularization is formulated as follows:
\begin{equation}
        \mathcal{L}_{\text{depth}} =\| Corr(d^{LN}, \hat{d}^{LN}) \|_{1}, 
\end{equation}
where
\begin{equation}
        Corr(d^{LN}, \hat{d}^{LN}) = \frac{\text{Cov}(d^{LN}, \hat{d}^{LN})}{\sqrt{\text{Var}(d^{LN}) \cdot \text{Var}(\hat{d}^{LN})}}
\end{equation}
refers to Pearson correlation, $\hat{d}^{LN}$ denotes the normalized depth patch rendered from the scene, and $d^{LN}$ represents the normalized depth patch generated by DPT.

\subsection{Loss Functions}
The total loss function is formulated as follows:
\begin{equation}
    \mathcal{L} = \omega_{0} \mathcal{L}_{\text{0}} + \omega_\text{{sem}} \mathcal{L}_{\text{sem}} + \omega_\text{{depth}} \mathcal{L}_{\text{depth}},
\end{equation}
where $\omega_{0}$, $\omega_\text{{sem}}$, and $\omega_\text{depth}$ are the weights, which are hyper-parameters.


\section{Experiments and Analysis}
\subsection{Experiment Details}
We conducted our experiments on the Local Light Field Fusion (LLFF) dataset \cite{mildenhall2019local}, which contains 8 scenes for training and testing, each comprising 20 to 62 images captured from different viewpoints. 
Following the configurations in \cite{niemeyer2022regnerf}, 
we select every eighth image for testing, and evenly sample three views from the remaining images for training.
We implemented our method on the images with 1/8 and 1/4 scales during training and testing. To implement the local depth regularization, we set the size of local patches extracted from the depth map to $126\times 126$. The number of iterations was fixed at $1.2\times 10^{4}$. All experiments were conducted on an NVIDIA RTX 4090 GPU. We evaluated model performance using Peak Signal-to-Noise Ratio (PSNR) and Structural Similarity Index Measure (SSIM) to assess reconstruction quality, and Learned Perceptual Image Patch Similarity (LPIPS) \cite{zhang2018unreasonable} to measure the perceptual quality of the rendered images.

\subsection{Experiments on LLFF Dataset}
We compare our method with other promising methods, including Mip-NeRF \cite{barron2021mip}, DietNeRF \cite{jain2021putting}, RegNeRF \cite{niemeyer2022regnerf}, FreeNeRF \cite{Yang2023FreeNeRF}, SparseNeRF \cite{wang2023sparsenerf}, 3DGS \cite{kerbl3Dgaussians}, and FSGS \cite{zhu2023FSGS}. The average performance of all methods is illustrated in Table~\ref{tb:LLFF}.

As observed, our method significantly outperforms SparseNeRF and FSGS by up to 0.95dB and 0.4dB, respectively. The LPIPS scores achieved by our method are also lower than all compared methods. These results reveal that our method achieves the best performance in terms of reconstruction and perceptual quality. In addition, the images rendered by different methods are shown in Fig.~\ref{fig:LLFF} for visual comparison. We can observe that NeRF-based methods, i.e., FreeNeRF \cite{Yang2023FreeNeRF} and RegNeRF \cite{niemeyer2022regnerf}, generate floaters in the scenes ``Leaves" and ``Horns".
Although FSGS shows a very satisfactory result, it 
loses some details in the scene ``Leaves", and has an incorrect prediction of edges in the scene ``Horns". Our method produces detailed results and correct structures, demonstrating semantic and geometric consistency.




\subsection{Ablation Studies}
In this section, we explore our proposed method in two aspects: (1) semantic regularization, and (2) local depth regularization. Quantitative results on the LLFF dataset under the 3-view setting are shown in Table \ref{tab:ablation}, and visual results are shown in Fig. \ref{fig:ablation}. Our proposed semantic regularization greatly improves quantitative metrics and the visual quality of rendered results, demonstrating the effectiveness of enhanced semantic consistency. Based on this, local depth regularization further improves the render quality with more accurate details. 

We also evaluate the sensitivity of loss weights of $\mathcal{L}_{\text{sem}}$ and $\mathcal{L}_{\text{depth}}$ by varying $\omega_\text{{sem}}$ and $\omega_\text{{depth}}$. Experimental results are shown in Fig. \ref{fig:parameters}. A smaller value of $\omega_\text{{sem}}$ (0.6) leads to greater fluctuations in PSNR, whereas an increase in $\omega_\text{{depth}}$ correlates with higher PSNR values.


\begin{table}[h]
    \centering
    \caption{
    Quantitative results of our SIDGaussian with (\textcolor{green}{\ding{51}}) or without (\textcolor{red}{\ding{55}}) proposed components.
    }
    \renewcommand\arraystretch{1.5}
    \begin{tabular}{@{}ccc c c c@{}}
        \toprule
        Semantic Regularization& Local Depth & \textbf{PSNR$\uparrow$} & \textbf{SSIM$\uparrow$} & \textbf{LPIPS$\downarrow$}\\ \midrule
        \textcolor{red}{\ding{55}} & \textcolor{red}{\ding{55}} & 20.31 & 0.652 & 0.288 \\
        \textcolor{green}{\ding{51}} & \textcolor{red}{\ding{55}} & 20.57 & 0.700 & 0.220 \\
        \textcolor{green}{\ding{51}} & \textcolor{green}{\ding{51}} & \textbf{20.71} & \textbf{0.708} & \textbf{0.205} \\ \bottomrule
    \end{tabular}
    \label{tab:ablation}
\end{table}

\begin{figure}[h]
    \centering
    \includegraphics[width=1.0\linewidth]
    {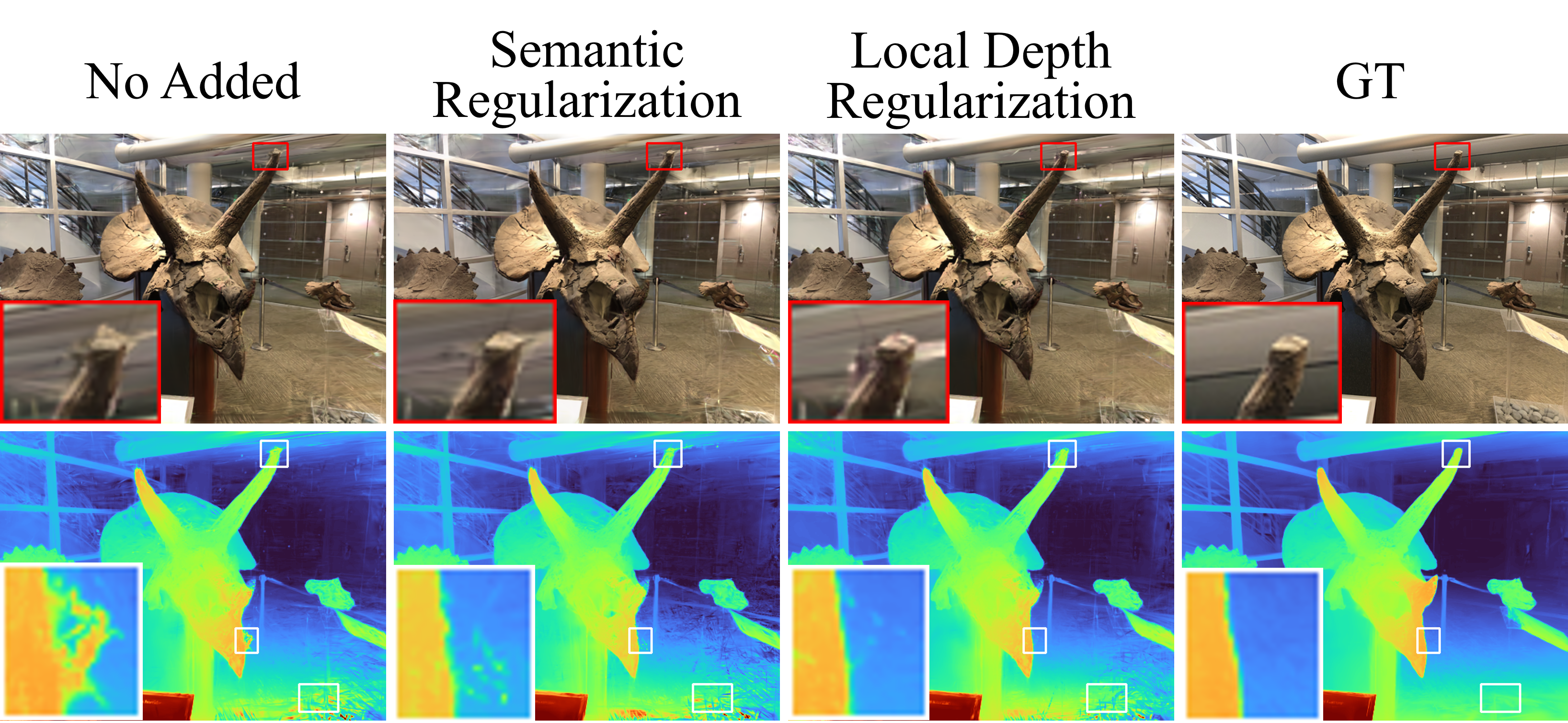}
    \caption{Visual results of our method with/without semantic regularization and local depth regularization.} 
    
    \label{fig:ablation}
\end{figure}

\begin{figure}[!ht]
    \centering
    \includegraphics[width=1\linewidth]{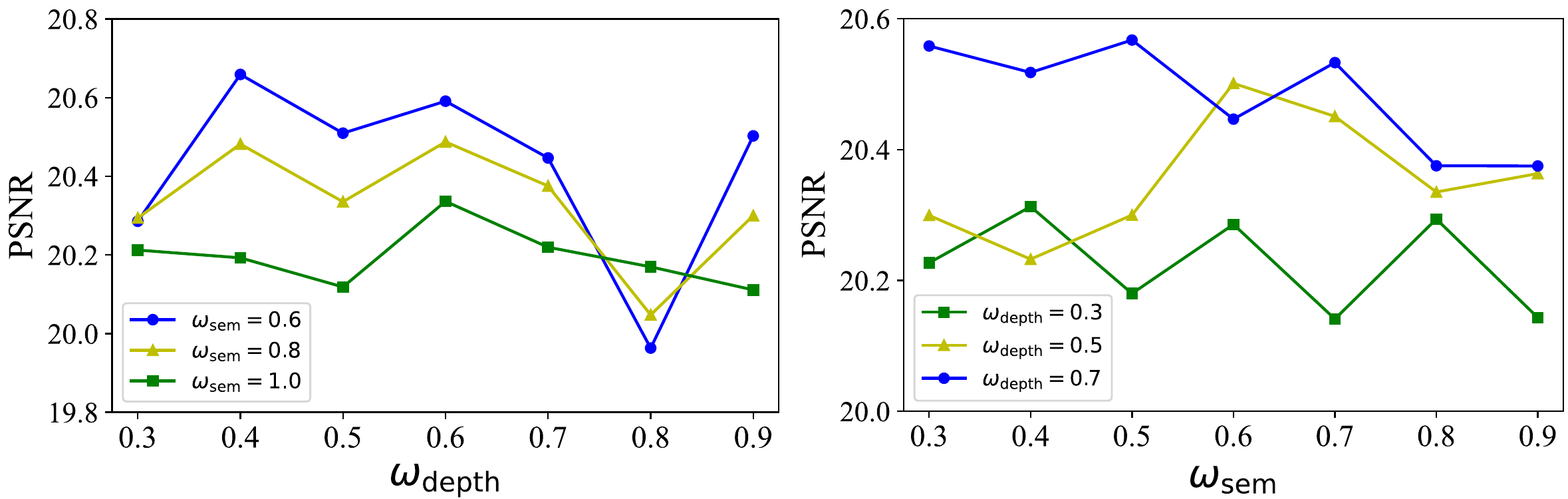}
    \caption{The influence of the weights $w_{\text{sem}}$ and $w_{\text{depth}}$ in terms of PSNR score. 
    }
    \label{fig:parameters}
\end{figure}


\section{Conclusion}

In this paper, we focus on novel view synthesis based on sparse inputs and propose a sparse-view 3D Gaussian splatting model, namely SIDGaussian. To ensure multi-view consistency, we propose a semantic regularization technique that aims to preserve the semantic coherence of rendered images across different viewpoints. Furthermore, we propose a local depth regularization to mitigate distorted content and enhance the detailed information of rendered images. The experiments demonstrate that our proposed SIDGaussian significantly outperforms other state-of-the-art methods in terms of PSNR and LPIPS on the LLFF dataset, achieving up to 0.4dB improvement. The images generated by our method exhibit the best visual quality with less distortion.

\bibliographystyle{ieeetr}
\bibliography{IEEEabrv}


\end{document}